\def\year{2018}\relax
\documentclass[letterpaper]{article} 
\usepackage{aaai18}  
\usepackage{times}  
\usepackage{helvet}  
\usepackage{courier}  
\usepackage{url}  
\usepackage{graphicx}  
\usepackage{subcaption} 
\usepackage[usenames, dvipsnames]{color}
\begin{document}
%
\title{Deception Detection in Videos}

\author{Zhe Wu$^1$ ~~~~~Bharat Singh$^1$~~~~~Larry S. Davis$^1$~~~~~V. S. Subrahmanian$^2$\\
$^1$University of Maryland ~~~~~~~~~~$^2$Dartmouth College\\
{\tt\small {\{zhewu,bharat,lsd\}@umiacs.umd.edu~~~vs@dartmouth.edu}
}
}


\maketitle
\begin{abstract}
We present a system for covert automated deception detection in real-life courtroom trial videos. We study the importance of different modalities like vision, audio and text for this task. On the vision side, our system uses classifiers trained on low level video features which predict human micro-expressions. We show that predictions of high-level micro-expressions can be used as features for deception prediction. Surprisingly, IDT (Improved Dense Trajectory) features which have been widely used for action recognition, are also very good at predicting deception in videos. We fuse the score of classifiers trained on IDT features and high-level micro-expressions to improve performance. MFCC (Mel-frequency Cepstral Coefficients) features from the audio domain also provide a significant boost in performance, while information from transcripts is not very beneficial for our system. Using various classifiers, our automated system obtains an AUC of 0.877 (10-fold cross-validation) when evaluated on subjects which were not part of the training set. Even though state-of-the-art methods use human annotations of micro-expressions for deception detection, our fully automated approach outperforms them by $5\%$. When combined with human annotations of micro-expressions, our AUC improves to 0.922. We also present results of a user-study to analyze how well do average humans perform on this task, what modalities they use for deception detection and how they perform if only one modality is accessible. Our project page can be found at \url{https://doubaibai.github.io/DARE/}.

\end{abstract}

\section{Introduction}
Deception is common in our daily lives. Some lies are harmless, while others may have severe consequences and can become an existential threat to society. For example, lying in a court may affect justice and let a guilty defendant go free. 
Therefore, accurate detection of a deception in a high stakes situation is crucial for personal and public safety. 

The ability of humans to detect deception is very limited. In ~\cite{bond2006accuracy}, it was reported that the average accuracy of detecting lies without special aids is $54\%$, which is only slightly better than chance. To detect deception more accurately, physiological methods have been developed. However, physiological methods such as the Polygraph, or more recent functional Magnetic Resonance Imaging (fMRI) based methods are not always correlated with deception ~\cite{farah2014functional}. Additionally, the cost of the equipment and the overt nature of the method make the utility of these devices limited for real-life deception detection.

Another line of work attempts to find behavioral cues for deception detection~\cite{depaulo2003cues}. These cues are faint behavioral residues, which are difficult for untrained people to detect. For example, according to ~\cite{ekman1969pan,ekman2009telling}, facial micro-expressions reflect emotions that subjects might want to hide. However, because of the variablity across different subjects, these micro-expressions are extremely difficult to detect using computer vision, especially in unconstrained settings.

 \begin{figure}[t!]
 \centering
 \includegraphics[width=1\linewidth]{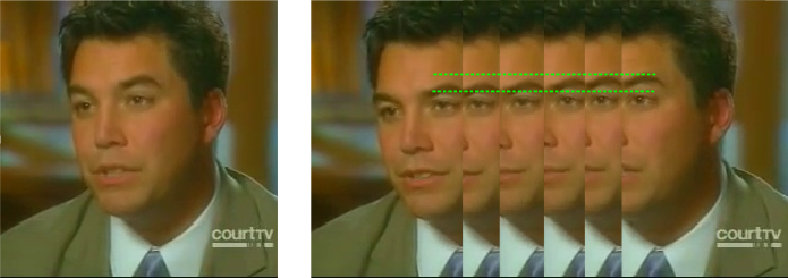}
\caption{Micro-expression: Eyebrows Raising. The left image and rightmost image in the image sequence are the same. However, it is much easier to detect the micro-expression in the image sequence when compared to the static image, as shown by the two green dotted lines.}
 \label{fig:change}
\end{figure}

It is usually much easier for humans to detect a subtle facial expression from videos than from static images \cite{grill1998cue}. For example, Fig.~\ref{fig:change} shows a comparison of static and dynamic representations of a simple micro micro-expression: Eyebrows Raise. Given only the left static image, people have a difficult time detecting that the eyebrows are raising. In contrast, we can clearly see from the right image sequence that the eyebrows are raising, even though the last image of the image flow is exactly the left static image.

Motivated by these observations, we propose to use motion dynamics for recognizing facial micro-expressions. 
This coincides with the psychological insights from ~\cite{duran2013exploring}, in which the authors suggest focusing on  dynamic motion signatures which are indicative of deception. 
To accomplish this, we design a two-level feature representation for capturing dynamic motion signatures. 
For the low-level feature representation, we use dense trajectories which represent motion and motion changes.
For the high-level representation, we train facial micro-expression detectors using low level features, and use their confidence score as high-level features. Experiments on 104 court room trial videos demonstrate the effectiveness and the complementary nature of our low-level and high-level features.

Deception is a complex human behavior where subjects try to inhibit their deceptive evidence, from facial expressions to gestures, from the way they talk to what they say. Thus, a reliable deception detection method should integrate information from more than one modality. Taking motivation from prior work ~\cite{perez2015deception,jaiswal2016truth}, we also include features from other modalities, specifically audio and text. These additional modalities improve AUC (Area under the precision-recall curve) of our automated system by ~$5\%$ to $0.877$. When using ground truth facial micro-expression annotations, the system obtains $0.922$ AUC which is $9\%$ better than the previous state-of-the-art. We also conduct user studies to analyze how well do average humans perform on this task, what modalities they use and how they perform if only one modality is accessible.

\section{Related Work}
Physiological measures have been considered to be useful in deception detection for a long time. Polygraph measures physiological indices such as blood pressure, heart rate, skin conductivity of the person under interrogation, but their reliability is questionable. Thermal imaging can record the thermal patterns~\cite{pavlidis2002human} and measure the blood flow of the body~\cite{buddharaju2005automatic}, but the technique requires expensive thermal cameras. Brain-based detectors, such as functional MRI, have recently been proposed to detect deception by scanning the brain and finding areas that are correlated with deception. Although it achieves high accuracy~\cite{kozel2005detecting,langleben2013using,farah2014functional}, important questions related to the working mechanism, reliability and the experimental setting are still open research problems ~\cite{langleben2013using,farah2014functional}. Furthermore, the above mentioned physiological measure based methods are overt and could be disrupted by the subject's counter preparation and behavior~\cite{ganis2011lying}. 

Among the covert systems, computer vision based methods play an important role. Early works ~\cite{lu2005blob,tsechpenakis2005hmm} used blob analysis to track head and hand movements, which were used to classify human behavior in videos in three different behavioral states. 
However, these methods used person specific sample images for training blob detectors, and since the database was small, the methods were prone to overfitting and did not generalize to new subjects. Based on Ekman's psychology research\cite{ekman1969pan,ekman2009telling} that some facial behaviors are involuntary and might serve as an evidence for deceit detection, several automatic vision-based systems were developed. Zhang et al. \cite{zhang2007real} tried to detect the differences between simulated facial expressions and involuntary facial expressions by identifying Deceit Indicators, which were defined by a group of specific Facial Action Units \cite{ekman1977facial}. However, this method requires people to manually label facial landmarks and input major components of FAU, thus is not fully automated. 
Besides, this method was only tested with static images, so essential motion patterns in facial expressions were not captured. \cite{michael2010motion} proposed a new feature called motion profiles to extend the head and hand blob analysis with facial micro-expressions. Although fully automatic, this method relies heavily on the performance of facial landmarks localization, and the experimental setting is very constrained. For unconstrained videos, the facial landmark localization method may be unreliable. 

These early works were mainly in low-stake situations. Recently, researchers focused more on high stake deception detection, so that experiments are closer to real life settings. 
In ~\cite{perez2015deception}, a new dataset containing real-life trial videos was introduced. 
In this work, a multi-modal approach is presented for the high-stake deception detection problem, but the method requires manual labelling of human micro-expressions. Furthermore, in the dataset, the number of videos for different trials varies a lot, biasing the results towards longer trials.
In \cite{su2016does}, the authors also collected a video database of high-stakes situations, and manually designed a variety of features for different facial parts. Again, the manually designed features require the facial landmarks to be accurately detected. Also, the method segmented each video into multiple temporal volumes and assumes all the temporal volume labels to be the same when learning the classifier. 
This could be incorrect for deception cases, because the deception evidence could be buried anywhere in the video. 

Deceptive behavior is very subtle and varies across different people. Thus, detecting these subtle micro motion patterns, e.g. micro facial expression,  itself is a challenging problem. In addition, Duran et al. ~\cite{duran2013exploring} suggested that research should focus more on the movement dynamics and behavior structure. Motivated by this, we directly describe behavioral dynamics without detecting facial landmarks, then use behavior dynamics to learn micro-expressions and deceptive behavior. We also include simple verbal and audio features, as other multi-modal approaches~\cite{perez2015deception,jaiswal2016truth}, into the overall covert automated system. 

\section{Method}
\begin{figure*}[ht!]
 \centering
 \includegraphics[width=0.9\linewidth]{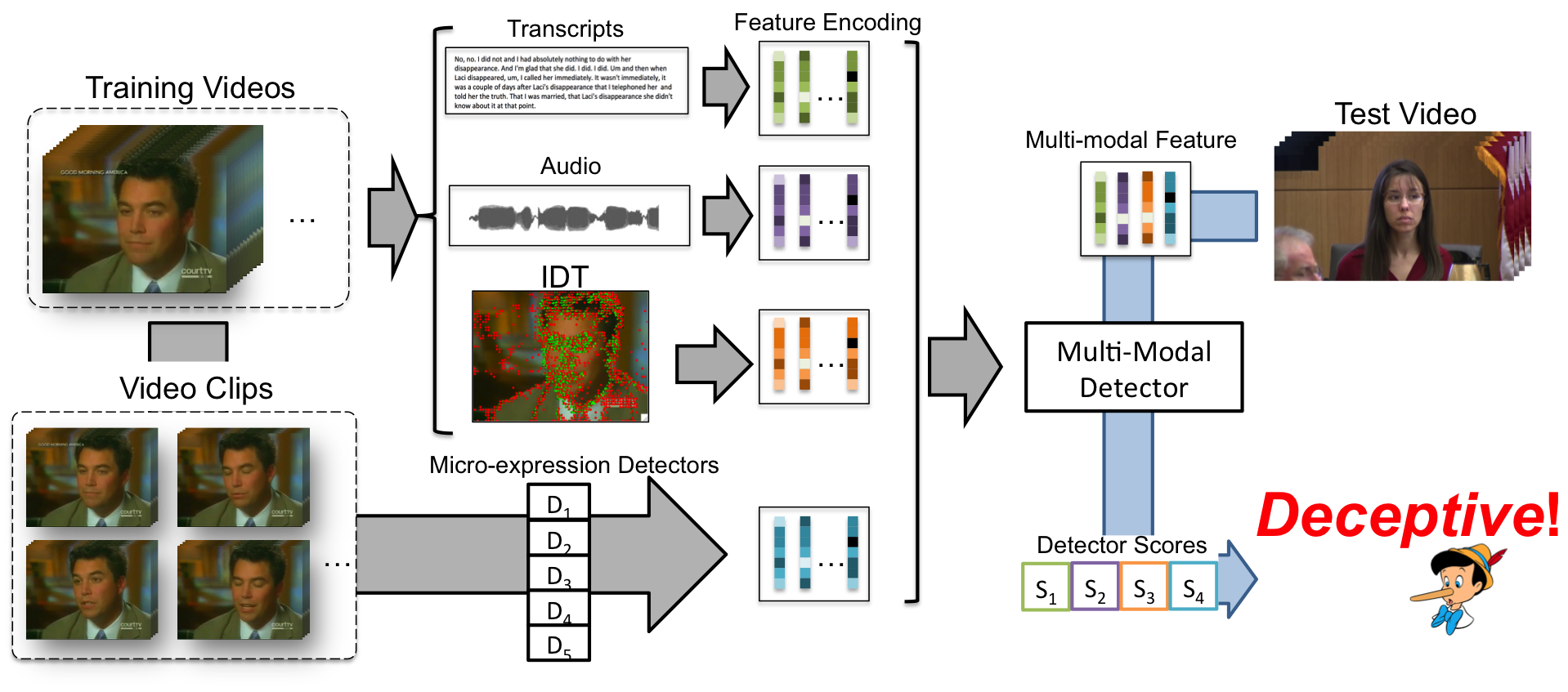}
\caption{Our automated deception detection framework.}
 \label{fig:framework}
\end{figure*}

Our automated deception detection framework consists of 3 steps: multi-modal feature extraction, feature encoding and classification. The framework is shown in Figure.~\ref{fig:framework}.

\subsection{Multi-Modal Feature Extraction}
\subsubsection{Motion Features}
Our input source are videos, where a person is making truthful or deceptive statements. The video acquisition conditions are unconstrained, so the subject's face may not always be viewed frontally or centered. Here, we employ IDT (Improved Dense Trajectory)~\cite{wang2016robust} features due to their excellent performance in action recognition, especially in unconstrained settings. 

IDT compute local feature correspondences in consecutive frames and estimate the camera motion using RANSAC. After cancelling the effect of camera motion, the method samples feature points densely at multiple spatial scales, then tracks them through a limited number of frames to prevent drifting. Within the space-time volume around the trajectory, the method computes HOG (histogram of oriented gradients), HOF (histogram of optical flow) \cite{laptev2008learning}, MBH (motion bountary histogram) \cite{dalal2006human} and trajectory descriptors. 
We found that the MBH descriptor works better than other descriptors (like HOG/HOF) for our task, because MBH computes optical flow derivatives and captures derivatives of motion rather than first order motion information. Since we want to detect micro-expressions, the descriptor should represent changes in motion rather than constant motion, which is captured in MBH.

\subsubsection{Audio Features}
We use MFCC (Mel-frequency Cepstral Coefficients) ~\cite{davis1980comparison} features as our audio features. MFCC has been widely used for ASR (Automatic Speech Recognition) tasks for over 30 years. We use the following MFCC extraction procedure: first estimate the periodogram of the power spectrum for each short frame, then warp to a Mel frequency scale, and finally compute the DCT of the log-Mel-spectrum. Then for each video, we have a series of MFCC features corresponding to short intervals. 
After MFCC features are extracted, we use GMM (Gaussian Mixture Model) to build an audio feature dictionary for all training videos. We treat all the audio features equally and use our feature encoding method to encode the whole sequence, similar to ~\cite{campbell2006support}. This is because we are not interested in speech content (spoken words), but in hidden cues of deception in the audio domain.

\subsubsection{Transcript Features}
For every video, we use Glove(Global Vectors for Word Representation)~\cite{pennington2014glove} for encoding the entire set of words in the video transcripts to one fixed-length vector. Glove is an unsupervised learning algorithm for representing words using vectors. It is trained using word co-occurrence statistics. As a result, the word vector representations capture meaningful semantic structure. Compared to other text-based deception detection methods~\cite{porter2010truth}, Glove is more widely applicable in unconstrained environments. 

We use the pre-trained Wikipedia 2014+ Gigaword5 corpus which contains 6 billion tokens in total. Each word is embedded in a 300 dimensional vector space. Again, we use GMM to learn a vocabulary for the word vectors and employ a Fisher Vector encoding, described below, to aggregate all the word vectors into a fixed-length representation for the entire transcript.

\begin{figure*}[ht!]
 \minipage{0.19\textwidth}
    \includegraphics[width=\linewidth,height=0.6\linewidth]{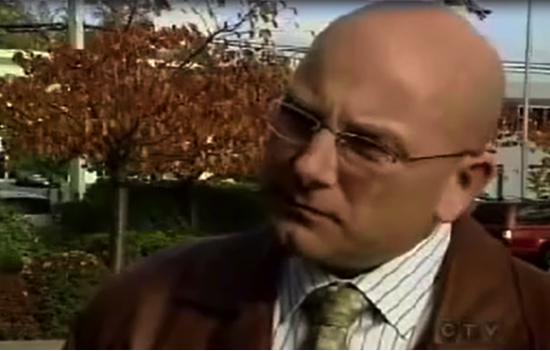}
    \label{fig:frown}
\endminipage\hfill
\minipage{0.19\textwidth}
    \includegraphics[width=\linewidth,height=0.6\linewidth]{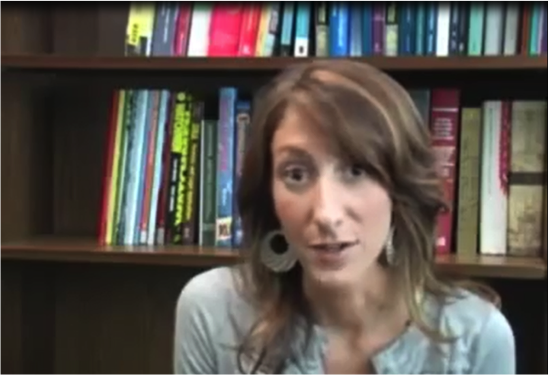}
    \label{fig:raise}
\endminipage\hfill
\minipage{0.19\textwidth}
    \includegraphics[width=\linewidth,height=0.6\linewidth]{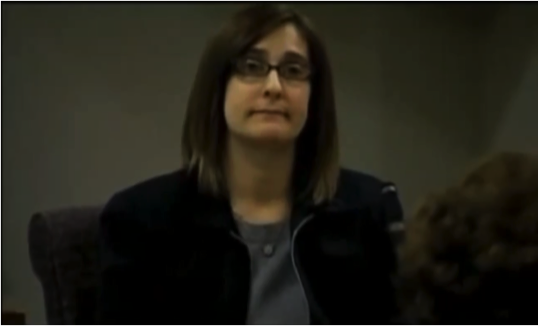}
    \label{fig:lipup}
\endminipage\hfill
\minipage{0.19\textwidth}
    \includegraphics[width=\linewidth,height=0.6\linewidth]{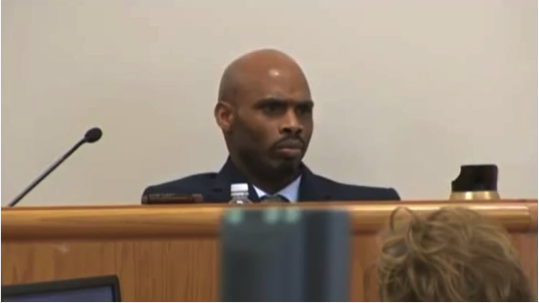}
    \label{fig:lippro}
\endminipage\hfill
\minipage{0.19\textwidth}
    \includegraphics[width=\linewidth,height=0.6\linewidth]{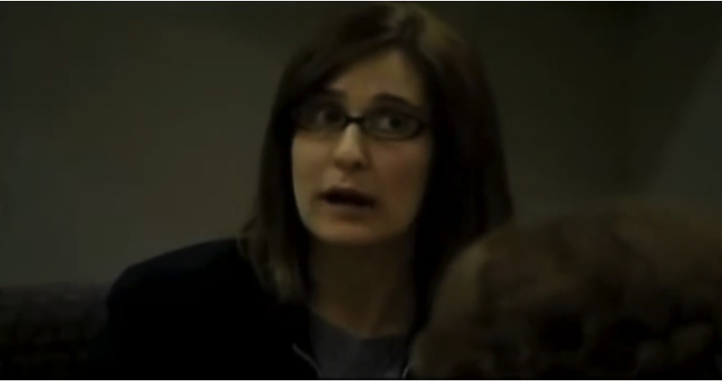}
    \label{fig:turn}
\endminipage\hfill 
\caption{Five most predictive micro-expressions, from left to right: Frowning, Eyebrows raising, Lip corners up, Lips protruded and Head Side Turn.}
\label{facialges}
\end{figure*}

\subsection{Feature Encoding}
Since the number of features is different for each video, we employ a Fisher Vector encoding to aggregate a variable number of features to a fixed-length vector. Fisher Vectors were first introduced in ~\cite{jaakkola1999exploiting} to combine the advantages of generative and discriminative models, and are widely used in other computer vision tasks, such as image classification~\cite{perronnin2007fisher}, action recognition~\cite{wang2016robust} and video retrieval~\cite{han2017vrfp}.

Fisher Vector encoding first builds a $K$-component GMM model $(\mu_i, \sigma_i, w_i:i=1,2,..., K)$ from training data, where $ \mu_i, \sigma_i, w_i$ are the mean, diagonal covariance, and mixture weights for the $i^{th}$ component, respectively. Given a bag of features $\{ x_1, x_2, ..., x_T\}$, its Fisher Vector is computed as:

\begin{equation}
\mathcal{G}_{{\mu}_i} = \frac{1}{T\sqrt{w_i}}\sum_{t=1}^{T} \gamma_t(i) \left(\frac{ x_t - \mu_i}{ \sigma_i} \right)
\end{equation}
\begin{equation}
\mathcal{G}_{{\sigma}_i} = \frac{1}{T\sqrt{2w_i}}\sum_{t=1}^{T} \gamma_t(i) \left( \frac{(x_t -  \mu_i)^2}{ \sigma_i^2} -  1 \right)
\end{equation}
where, $\gamma_t(i)$ is the posterior probability. Then all the $\mathcal{G}_{{\mu}_i}$ and $\mathcal{G}_{{\sigma}_i}$ are stacked to form the $2DK$-dimension Fisher Vector, where $D$ is the dimensionality of local feature $x_t$. 

\subsection{Facial Micro-Expression Prediction}
The multi-modal features discussed above are low-level features. 
Here, we introduce the high-level features used to represent facial micro-expressions. According to \cite{ekman1969pan,ekman2009telling}, facial micro-expressions play an important role in predicting deceptive behavior. To investigate their effectiveness, \cite{perez2015deception} manually annotated facial expressions and use binary features derived from the ground truth annotations to predict deception. They showed the five most predictive micro-expressions are: Frowning, Eyebrows raising, Lip corners up, Lips protruded and Head Side Turn. Samples of these facial micro-expressions are shown in Figure.~\ref{facialges}.

We use the low-level visual features to train micro-expression detectors, and then use the predicted scores of the micro-expression detectors as high-level features for predicting deception. We divide each video in the database into short fixed-duration video clips and annotate these clips with micro-expression labels. Formally, given a training video set $V = \{v_1, v_2,..., v_N\}$, by dividing each video into clips, we obtain a training  set $C = \{v_i^j\}$. The annotation set is  $L=\{l_i^j\}$, $i\in[1,N]$ denotes the video id, the superscript $j\in[1,n_i]$ denotes the clip id
, $n_i$ is the number of clips for video $i$ and the duration of $v_i^j$ is a constant (4 seconds in our implementation). The dimension of $l_i^j$ is the number of micro-expressions. Then we train a set of micro-expression classifiers using the clips $C$, and apply the classifiers on test video clips $\tilde{C}$ to generate the predicted score $\tilde{L} = \{\tilde{l}_i^j\}$. These score vectors are pooled by averaging them over all clips in a video to produce a video score vector.

\subsection{Deception Detection}
The Fisher vector encoding of the low level features and the video level score vector are then used to train four binary deception classifiers. Three of those classifiers are based on the visual, auditory and text channels for which GMM's were constructed, and the fourth uses the pooled score vectors for the micro-expression detectors. Denote the prediction score of the multi-modal Fisher Vector and the high-level micro-expression feature as $\{S_{m_i}\}, i\in[1,3]$ and $S_{high}$. The final deception score $S$ is obtained by late fusion, given by:
\begin{equation}
S = \sum_{i} {\alpha_iS_{m_i}} + \alpha_{high}S_{high}
\end{equation}
where $\alpha_i, \alpha_{high} > 0$ and $\sum_{i=1}^4\alpha_i + \alpha_{high} = 1$. The values of  $\alpha_i$ and $\alpha_{high}$ are obtained by cross validation.

\section{Experiments and Results}
\subsection{Dataset}
\begin{figure}[t!]
 \centering
 \includegraphics[width=1\linewidth]{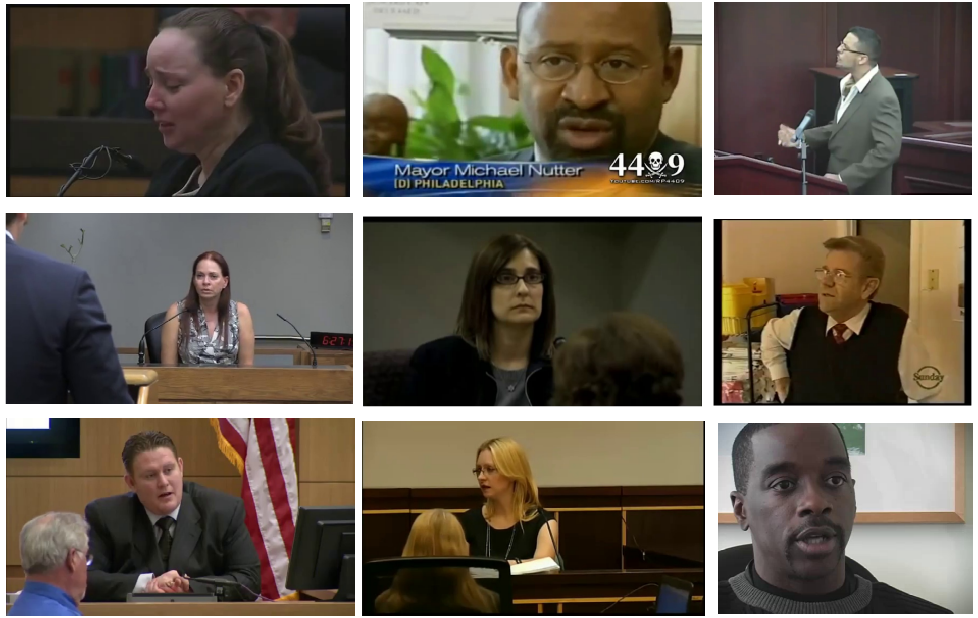}
\caption{Video Samples in the Real-life Trial Deception Database depicting the variation in human face size, pose and illumination.}
 \label{fig:dataset}
\end{figure}

We evaluate our automated deception detection approach on a real-life deception detection database~\cite{perez2015deception}. This database consists of 121 court room trial video clips.
Videos in this trial database are unconstrained videos from the web. Thus, we need to handle differences in the viewing angle of the person, variation in video quality and background noise, as shown in Figure.~\ref{fig:dataset}. 

We use a subset of 104 videos from the trial database of 121 videos, including 50 truthful videos and 54 deceptive videos. The pruned videos have either significant scene change or human editing. In the experiments shown below, we do not report the results, as described  in ~\cite{perez2015deception}. Instead, we re-implement the method (referred to Ground Truth micro-expressions) on our training and test splits to avoid over fitting to identities rather than deception clues.

The dataset contains only 58 identities, which is less than the number of videos and often the same identity is either uniformly deceptive or truthful. This means a deception detection method may simply degenerate to  person re-idetification, if videos of the same person were included in both the training and test splits. To avoid this problem, we perform 10-fold cross validation using identities instead of video samples for all the following experiments, i.e. no identity in the test set belongs to the training set.

\begin{table}[t!]
\centering
\begin{tabular}{|c|c|}
 \hline
 Micro-Expression & IDT+FV\\
 \hline\hline
 Eyebrows Frown& 0.6437 \\
 Eyebrows Raise & 0.6633\\
 Lips Up & 0.4791\\
 Lips Protruded & 0.7512\\
 Head Side Turn & 0.7180\\
 \hline
 Mean & 0.6511\\
 \hline
\end{tabular}
\caption{Micro-Expression Detector Performance}
\label{tab:ges}
\end{table}

\begin{table*}[h]
\centering
\begin{tabular}{|c|c|c|c|c|c|c|c|c|}
\hline
Features & L-SVM & K-SVM & NB & DT & RF & LR & Adaboost\\
\hline\hline
IDT & 0.7731 & 0.6374 & 0.5984 & 0.5895 & 0.5567 & 0.6425 & 0.6591\\
\hline
MicroExpression & 0.7502 & 0.7540 & 0.7629 & 0.7269 & 0.8064 & 0.7398 & 0.7507\\
\hline
Transcript & 0.6457 & 0.4667 & 0.6625 & 0.5251 & 0.6172 & 0.5643 & 0.6416\\
\hline
MFCC & 0.7694 & 0.8171 & 0.6726 & 0.4369 & 0.7393 & 0.6683 & 0.6900\\
\hline\hline
IDT+MicroExpression & 0.8347 & 0.7540 & 0.7629 & 0.7687 & 0.8184 & 0.7419 & 0.7507\\
\hline
IDT+MicroExpression+Transcripts & 0.8347 & 0.7540 & 0.7776 & 0.7777 & 0.8184 & 0.7419 & 0.7507\\
\hline
IDT+MicroExpression+MFCC & 0.8596 & 0.8233 & 0.7629 & 0.7687 & 0.8477 & 0.7894 & 0.7899 \\
\hline
All Modalities & \textbf{0.8773} & 0.8233 & 0.7776 & 0.7777 & 0.8477 & 0.7894 & 0.7899\\
\hline
\end{tabular}
\caption{Deception Detection results using different feature and classifier combinations. First 4 rows are results of independent features. Last 4 rows are late fusion results of multi-modal features.}
\label{tab:res}
\end{table*}

\begin{table*}[h]
\centering
\begin{tabular}{|c|c|c|c|c|c|c|c|c|}
\hline
Features & L-SVM & K-SVM & NB & DT & RF & LR & Adaboost\\
\hline\hline
GTMicroExpression & 0.7964 & 0.8102 & 0.8325 & 0.7731 & 0.8151 & 0.8275 & 0.8270\\
\hline
GTMicroExpression+IDT & 0.8456 & 0.8137 & 0.8468 & 0.7834 & 0.8205 & 0.8988 & 0.8270\\
\hline
GTMicroExpression+IDT+Transcript & 0.8594 & 0.8137 & 0.8923 & 0.8074 & 0.8205 & 0.8988 & 0.8270\\
\hline
GTMicroExpression+IDT+MFCC & 0.8969 & 0.9002 & 0.8668 & 0.7834 & 0.8319 & 0.9221 & 0.8320\\
\hline
GTMicroExpression+All Modalities& 0.9065 & 0.9002 & 0.8905 & 0.8074 & 0.8731 & \textbf{0.9221} & 0.8321\\
\hline
\end{tabular}
\caption{Deception Detection results with Ground Truth micro-expression features and other feature modalities.}
\label{tab:resGT}
\end{table*}

\subsection{Micro-Expression Prediction}
We first analyze the performance our micro-expression prediction module. We sample frames for each video clip using a frame rate of 15 fps. The motion features are Improved Dense Trajectories which are represented with a Fisher Vector encoding. The micro-expression detectors are trained using a linear kernel SVM using LibSVM~\cite{chang2011libsvm}. The results are shown in Table.~\ref{tab:ges}, and we report AUC (Area under the precision-recall curve). We will show in the following experiments that even though an AUC of 0.6511 is not high for detecting micro-expressions, the high-level features representing the probability of micro-expressions still provide good performance on the final deception detection task. We believe deep learning based approaches could perform better at predicting micro-expressions; however, with the limited amount of training data available in this dataset, it is problematic to train such techniques. We did experiment with off-the-shelf CNN features for classifying micro-expressions, but their performance was significantly worse than IDT.


\subsection{Deception Detection}
We now evaluate our automated deception detection system. We first test four individual features: IDT (Improved Debse Trajectory), high-level micro-expression scores, verbal features and MFCC audio features. Then we test different combinations of multi-modal features. To test the reliability and robustness of the features, we use several widely used binary classifiers in our experiments, which are Linear SVM, Kernel SVM, Naive Bayes, Decision Trees, Random Forests, Logistic Regression and Adaboost. We use a polynomial kernel for Kernel SVM because it performs best. For Naive Bayes classifier, we use normal distributions and remove the feature dimensions which have zero variance before fitting. For logistic regression, we use Binomial distribution. In Random Forest, the number of trees is 50. 
In Adaboost, we use decision trees as the weak learners. All experiments are conducted using 10-fold cross validation across different feature sets and classifiers. 

The results, measured by AUC, are shown in Table.~\ref{tab:res}. The first 4 rows are results from one modality, while the last 4 rows are after late fusion of multi-modal features. Each column corresponds to one type of classifier. We can see the highest AUC (0.8773) is after late fusion, which uses all modality features and a linear SVM classifier. This performance is much better than using Ground Truth micro-expression features (0.8325). 

\subsubsection{Performance of Different Classifiers}
SVM and Random Forest perform better compared to other classifiers like Naive Bayes and Logistic Regression because they are discriminative. One interesting finding is that different classifiers are good at utilizing different feature modalities. For example, we observe that linear SVM works best on IDT features, Random Forest works best on high-level micro-expression features and Kernel SVM performs best on MFCC features. However, when we aggregate multi-modal features using late fusion, the performance of different classifiers converges.

\subsubsection{Performance of Different Modalities}
We observe that IDT features obtain an AUC of 0.7731. Although the proposed high-level micro-expression features could not accurately predict micro-expressions, they help in improving deception detection. MFCC features obtain the highest AUC using a single modality, showing the importance of audio features in the deception detection task. 
The transcript feature obtains the lowest performance, mainly because the pre-trained word vector representation does not capture the underlying complicated verbal deception cues. Nevertheless, the entire system still benefits  from the transcript features after late fusion.

\subsubsection{Analysis of Late Fusion}
Although the performance of different modalities are different for each classifier, the overall performance improves when we combine different modalities. Combining the scores of the classifier trained on IDT features with the classifier trained on micro-expression predictions helps us obtain an AUC of 0.8347, which is the performance of visual modality. Thus, even though the micro-expression detectors are trained using IDT features, the low-level and high-level classifiers are complementary. Other modalities like text and audio improve performance by $4\%$ for the overall system.

\subsection{Deception Detection with Ground Truth Micro-Expressions}
Since the high-level feature is the prediction score of trained micro-expression detectors, one interesting question is how the performance will be affected if we use the Ground Truth micro-expression features, as in ~\cite{perez2015deception}. In the following experiment, we use the GT micro-expression feature as the baseline, and test how the performance changes with other feature modalities. Table.~\ref{tab:resGT} shows the results, measured by AUC. Note that we re-ran this study because we do not use the same identity in the training and test splits.

From Table.~\ref{tab:res}, we observe that the performance of GTMicroExpression~\cite{perez2015deception} alone is better than high-level micro-expression features (which is the confidence score of the micro-expression classifier). With the addition of IDT features, our vision system improves by more than $5\%$ (0.8988 AUC). This proves the effectiveness of motion-based features. After late fusion with results of transcript and MFCC features, the performance of the overall system is 0.9221 AUC, which is better than the proposed fully automated system. This suggests that developing more accurate methods for detecting micro-expressions is a potential direction for improving deception detection in the future.

\begin{figure}[h!]
\begin{subfigure}{.47\textwidth}
  \centering
  \includegraphics[width=0.95\linewidth]{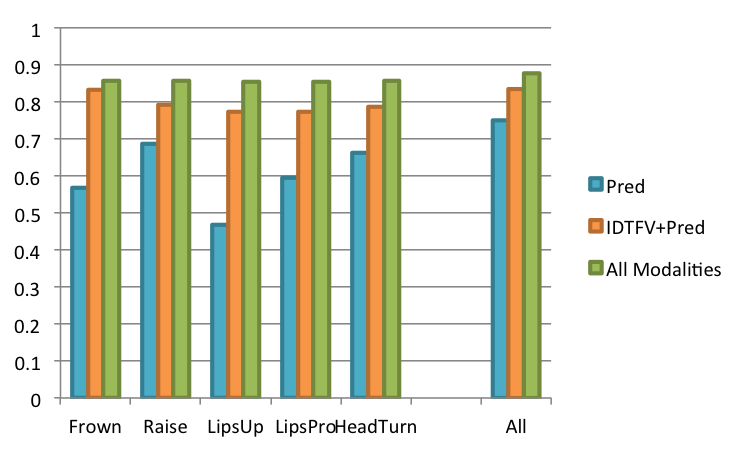}
  \caption{Predicted Micro-Expressions}
  \label{fig:unary_pred}
\end{subfigure}%
\\
\begin{subfigure}{.47\textwidth}
  \centering
  \includegraphics[width=0.95\linewidth]{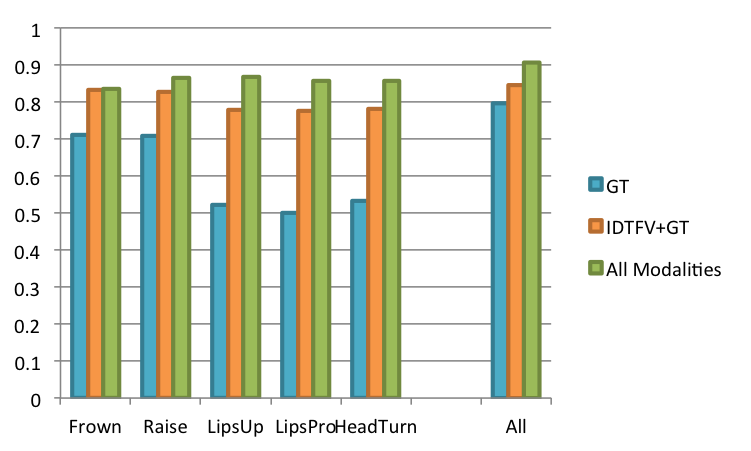}
  \caption{Ground Truth Micro-Expressions}
  \label{fig:unary_gt}
\end{subfigure}
\caption{Analysis of predicted micro-expressions and Ground truth micro-expressions. Each group depicts the performance with only that micro-expression. The performance of using all micro-expressions is shown in the rightmost group.}
\label{fig:unary}
\end{figure}

\subsection{Analysis of Micro-Expressions}
We investigate the effectiveness of each individual micro-expression. 

For each micro-expression, we test the performance by using the high-level micro-expression score feature, with low level motion features and other modalities, shown in Figure.~\ref{fig:unary}. 
The performance of using all micro-expressions is shown for comparison. We use linear SVM as the classifier in this study, as it was the best individual classifier. 
From Figure.~\ref{fig:unary}, we observe that ``Eyebrows Raise'' is more effective than other micro-expressions in both predicted micro-expressions and ground truth micro-expressions. ``Head Side Turn'' is also helpful when using predicted micro-expressions, see Figure.~\ref{fig:unary_pred}. This is different from results obtained from ground truth micro-expressions.
On the other hand, ``Frown'' works better using ground truth feature than predicted feature, possibly because the ``Frown'' detector is not accurate enough, as also suggested in Table.~\ref{tab:ges}.

\subsection{User Study}
To test human performance on this task, we perform user studies using AMT (Amazon Mechanical Turk). First, we ask 10 different people to watch each video and decide if they consider the subject in the video truthful or not. Each annotator is assigned 5 videos of 5 different identities to ensure no identity specific bias is used for deceit prediction. We also record if image, audio or transcripts were helpful for their decision. Note that here the decision is made using all modalities. The percentage of votes per video is used as the score for deception. The AUC for human prediction is 0.8102. This shows that this dataset is relatively easier than previous studies where predictions of people on this task were almost chance. Nevertheless, even on this dataset, it is not obvious if the subject is deceptive or not. 

\begin{figure}[t!]
 \centering
 \includegraphics[width=0.9\linewidth]{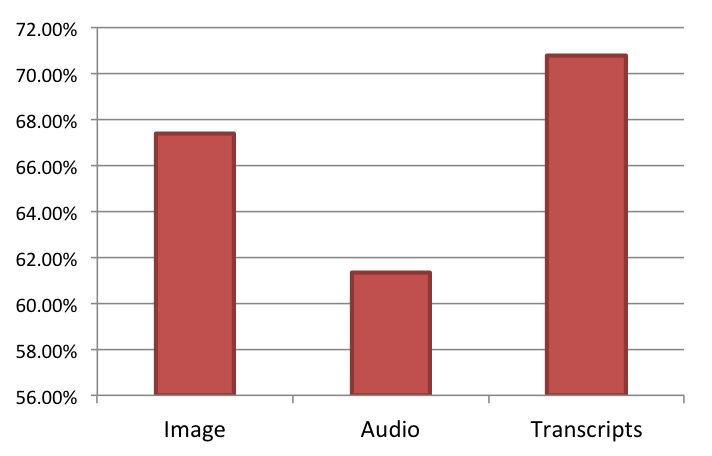}
\caption{The importance of modalities for humans in making decisions.}
 \label{fig:AMT}
\end{figure}

When making the decision, $67.4\%$ of the time users rely on visual cues, $61.3\%$ of the time on audio, and $70.7\%$ of the time on transcripts, as shown in Figure.~\ref{fig:AMT}. Note that for each video, people can select multiple modalities as helpful. From this data, we notice that people tend to make decisions based on spoken content, as this is a semantic-level feature. 
Only half of the people think that audio helps them making decision, while in our system, audio features are very effective. Therefore, we conducted another user study where we only show one modality at a time to each individual user because when multiple sources of information are available simultaneously, it is not easy to tell which source was helpful for making the final decision.

\begin{figure}[t!]
 \centering
 \includegraphics[width=1\linewidth]{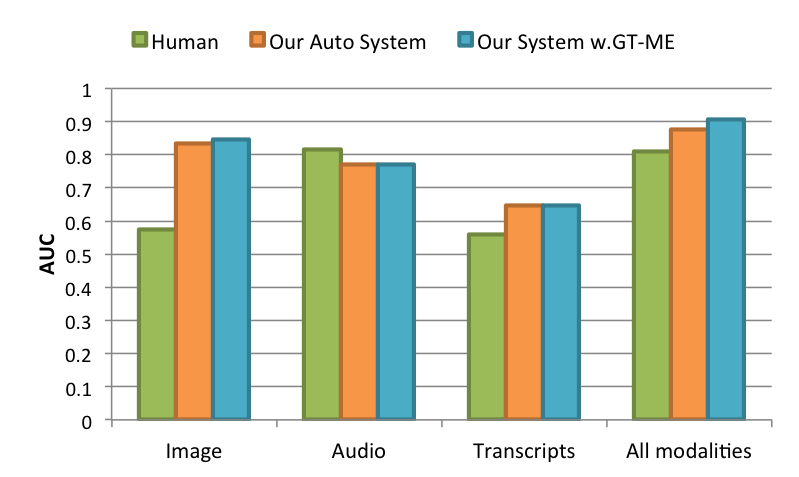}
\caption{Human performance in deception detection using different modalities is compared with our automated system and our system with Ground Truth micro-expressions.}
 \label{fig:AMT2}
\end{figure}

To test human performance on each modality, we ask 5 people to watch each video without sound, 5 people to listen to the audio and 5 people to read the transcripts of the video. Therefore in each study, subjects have access to one modality only. Note that the same person is not shown any other modality from the same video again after being shown one modality. 
The results shown in Figure.~\ref{fig:AMT2} are of our system using linear SVM as classifier along with human performance. We can see that with only visual modality, there is a huge performance gap between human performance and our system. This shows that although humans lack the ability of predicting deceptive behavior with visual cues alone, our computer vision based system is significantly better.
On the other hand, with only audio, human performance is as good as when all modalities are accessible. But when only transcripts of videos are provided, the performance drops significantly both for humans and our system. This suggests that audio information plays an essential role for humans for predicting deceptive behavior, while transcripts are not that beneficial. 
With all modalities, our automated system is about $7\%$ better compared to an average person, while our system with Ground Truth micro-expressions is about $11\%$ better. 

For future work, we believe better models for representing audio would be a promising direction for improving performance of our system. In addition, designing hybrid human-computer systems might be very important for developing a robust and accurate deception detection system.

\section{Conclusion}
A system for covert automatic deception detection using multi-modal information in a video was presented. We demonstrated that deception can be predicted independent of the identity of the person. Our vision system, which uses both high-level and low level visual features, is significantly better at predicting deception compared to humans. When complementary information from audio and transcripts is provided, deception prediction can be further improved. These claims are true over a variety of classifiers verifying the robustness of our system. To understand how humans predict deception using individual modalities, results of a user study were also presented. As part of future work, we believe collecting more data for this task would be fruitful as more powerful deep learning techniques can be employed. Predicting deception in a multi-agent setting using information available in video would be a promising future direction as the system would need to understand the conversation between identities over time and then arrive at a logical conclusion.

\section{Acknowlegement}
This research was funded by ARO Grant W911NF1610342. We would like to thank Srijan Kumar for helpful discussions during this project.

\bibliographystyle{aaai}\bibliography{reference.bib}
\end{document}